\documentclass{article}

\usepackage{arxiv}  
\usepackage[utf8]{inputenc} % allow utf-8 input
\usepackage[T1]{fontenc}    % use 8-bit T1 fonts
\usepackage{hyperref}       % hyperlinks
\usepackage{url}            % simple URL typesetting
\usepackage{booktabs}       % professional-quality tables
\usepackage{amsfonts}       % blackboard math symbols
\usepackage{nicefrac}       % compact symbols for 1/2, etc.
\usepackage{microtype}      % microtypography
\usepackage{xcolor}         % colors
\usepackage{algorithm}

\usepackage{algorithmic}
\usepackage{amsmath, amsthm, amssymb,amsfonts}
\usepackage{hyperref}  
\usepackage{graphicx,psfrag,epsf}
\usepackage{enumerate}
\usepackage{bm}
\usepackage{titletoc}

\titlecontents{section}
    [0em] % Left margin
    {\vspace{.5em}\bfseries} % Formatting
    {\contentslabel{1.5em}} % Numbering
    {} % Before-entry text
    {\titlerule*[.5em]{.}\contentspage} % Filler and page

\usepackage{listings}
\usepackage{color}

\definecolor{dkgreen}{rgb}{0,0.6,0}
\definecolor{gray}{rgb}{0.5,0.5,0.5}
\definecolor{mauve}{rgb}{0.58,0,0.82}

\usepackage[capitalize]{cleveref}
\crefname{section}{Sec.}{Secs.}
\Crefname{section}{Section}{Sections}
\Crefname{table}{Table}{Tables}
\crefname{table}{Tab.}{Tabs.}

\arxivtitlerunning{Probabilistic Skip Connections}

\begin{document}
\twocolumn[
\arxivtitle{Probabilistic Skip Connections for Deterministic Uncertainty Quantification in Deep Neural Networks}
\arxivsetsymbol{equal}{*}%

\begin{arxivauthorlist}
\arxivauthor{Felix Jimenez}{wisc}
\arxivauthor{Matthias Katzfuss}{wisc}
\end{arxivauthorlist}

\arxivaffiliation{wisc}{Department of Statistics, University of Wisconsin--Madison}

\arxivcorrespondingauthor{Felix Jimenez}{felix.m.jimenez3@gmail.com}
\arxivkeywords{Machine Learning}

\vskip 0.3in
]
\printAffiliationsAndNotice{}

\begin{abstract}
Deterministic uncertainty quantification (UQ) in deep learning aims to estimate uncertainty with a single pass through a network by leveraging outputs from the network's feature extractor. Existing methods require that the feature extractor be both \textit{sensitive} and \textit{smooth}, ensuring meaningful input changes produce meaningful changes in feature vectors. Smoothness enables generalization, while sensitivity prevents \textit{feature collapse}, where distinct inputs are mapped to identical feature vectors. To meet these requirements, current deterministic methods often retrain networks with spectral normalization. Instead of modifying training, we propose using measures of \textit{neural collapse} to identify an existing intermediate layer that is both \textit{sensitive} and \textit{smooth}. We then fit a probabilistic model to the feature vector of this intermediate layer, which we call a probabilistic skip connection (PSC). Through empirical analysis, we explore the impact of spectral normalization on neural collapse and demonstrate that PSCs can effectively disentangle aleatoric and epistemic uncertainty. Additionally, we show that PSCs achieve uncertainty quantification and out-of-distribution (OOD) detection performance that matches or exceeds existing single-pass methods requiring training modifications. By retrofitting existing models, PSCs enable high-quality UQ and OOD capabilities without retraining.
\end{abstract}

\section{Introduction}
Uncertainty quantification (UQ) for deep learning has traditionally relied on neural network ensembles \citep{lakshminarayanan2017deepensembles} or Bayesian methods that sample network weights with techniques like Markov chain Monte Carlo (MCMC) \citep{neal2012bayesiandl}, Monte Carlo dropout \citep{gal2016dropout} or other methods approximating the posterior for network weights \citep{maddox2019swag,ritter2018kfac}. These methods involve either training the network multiple times or performing several passes through a single network, both of which are computationally intensive. 

\begin{figure}[ht]
    \centering
    \includegraphics[width=1.0\linewidth]{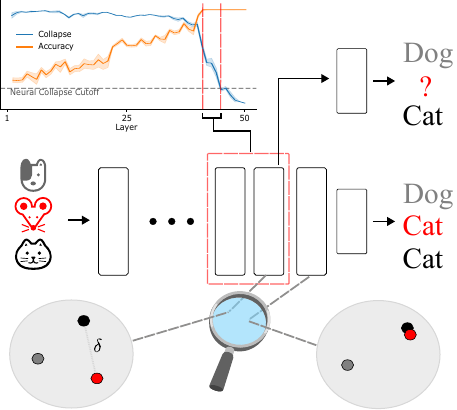}\hfill
    \caption{\textbf{Neural-collapse metrics help identify intermediate layers without feature collapse but high nearest-centroid accuracy, allowing us to place probabilistic skip connections (PSCs) that enhance uncertainty quantification.} The top left shows the trade-off between collapse and accuracy across network depth. In this example, dog, mouse, and cat images are processed through a network trained only on dog and cat images, with the mouse being out-of-distribution (OOD). Without a PSC, the network incorrectly labels the mouse as ``Cat". However, adding a PSC at a layer with high accuracy and low collapse gives a model that honestly reflects the uncertainty for the mouse image and correctly labels the two in-distribution instances. The magnifying glass view illustrates why: in pre-collapse layers, the feature vector difference, $\delta$, between the cat and mouse remains large. By the penultimate layer, this difference collapses, leading to overconfidence. The crux is to identify an appropriate intermediate layer that maintains feature sensitivity while preserving predictive performance.}
    \label{fig:visual_abstract}
\end{figure}

To address this burden, single-pass deterministic UQ methods have emerged as a promising solution \citep{amersfoot2020DUQ,liu2020sngp,mukhoti2023ddu} and estimate uncertainty with one forward pass through the network by using the network's feature maps. However, if the network is experiencing \textit{feature collapse}, a phenomenon where unique inputs are mapped to identical feature maps, then such methods will not work. Therefore, methods using feature maps rely on well structured feature spaces and distance-aware output layers, requiring modified network training and a particular network architecture.

An alternative approach is to use the outputs of intermediate layers where feature collapse has not occurred \cite{jimenez2023dve}, but this work has been limited to regression. As of yet, there has not been a single-pass deterministic uncertainty method that can be dropped into an existing classification neural network that does not require changing the network or its training procedure. 

In this work, we propose probabilistic skip connections (PSCs), which allow for deterministic UQ by identifying the intermediate layers of a neural network that generate feature maps that can be used with existing deterministic UQ methods. We find these layers by measuring neural collapse, a phenomenon closely related to but distinct from feature collapse, and show that our procedure allows for deterministic UQ in the presence of feature collapse, even for networks for which existing deterministic UQ methods were not applicable.  An overview of our approach is shown in Figure \cref{fig:visual_abstract}. 

Our contributions can be summarized as follows: 
\begin{itemize}
    \item We propose PSCs to allow for deterministic UQ for classification networks without needing to retrain the feature extractor with spectral normalization.
    \item Through an empirical study, we motivate our constructions of PSCs by examining the impact of spectral normalization on intermediate neural collapse.
    \item We show our approach rivals and often surpasses deterministic UQ methods, which require retraining, in both OOD detection and in-distribution UQ performance.
    \item We demonstrate that deterministic UQ methods extend to a larger family of neural networks than previously suggested.
\end{itemize}

\section{Background}\label{sec:background}
In this section, we provide a brief review of UQ, cover essential topics from feature geometry in deep learning, and discuss how feature collapse presents a challenge to performing deterministic UQ for deep learning. 

\textbf{Notation and terminology} for the remainder of the paper is described below. This work assumes neural networks of the form, $f(\bm x) = g \circ h (\bm x)$, where $h$ is referred to as the feature extractor and $g$ is the output layer. The function $h$ is assumed to be the composition of $L$ layers, $h(\bm x) = h_L\circ ... \circ h_1(\bm x)$, and we denote the image of $h$ as $\mathcal{H}$, which we refer to as the \textit{feature space}. Points in the feature space, i.e., the outputs of the feature extractor, are called feature maps. Then $g$ is taken to be a model mapping the feature space to logit space.

For input $\bm x \in \mathcal{X}$ and hidden layer $j \in \{1,...,L\}$, we refer to $h^{(j)}(\bm x) := h_j \circ ... \circ h_1(\bm x)$ as an intermediate representation of $\bm x$. The final intermediate representation, $h(\bm x) = h_L\circ ... \circ h_1(\bm x)$, is referred to as the feature map of $\bm x$. 

Let $d_{\mathcal{X}}$ denote an appropriate distance metric on the data manifold, such that $d_{\mathcal{X}}(\bm x_1, \bm x_2) > 0$ implies that $\bm x_1, \bm x_2 \in \mathcal{X}$ are semantically distinct. We abbreviate in-distribution as iD and out-of-distribution as OOD.

The \textit{spectral norm} of a matrix, $\bm A \in \mathbb{R}^{n\times m}$, is given by $\sigma_{\max}(\bm A)$, where $\sigma_{\max}$ denotes the matrix's largest singular value. \textit{Spectral normalization} (SN) \citep{miyato2018SN} is a technique that modifies neural network training by normalizing the weight matrices of network layers with their spectral norm throughout training.

\subsection{Uncertainty quantification}
We now introduce the framework we use for UQ in supervised learning. Consider a conditional distribution $p(y | \bm x)$, where $\bm x \in \mathcal{X} \subset \mathbb{R}^d$ represents inputs, and $y \in \mathcal{Y}$ represents corresponding responses. Given a dataset $\mathcal{D} = \{\bm x_i, y_i\}$, we learn a parametric function $f_{\bm \theta}: \mathcal{X} \rightarrow \mathcal{Y}$ by minimizing the empirical risk $\sum_{i=1}^n \mathcal{L}\left(f_{\bm \theta}(\bm x_i), y_i)\right)$. 

For a new input, $\bm x^*$, we use the tuned model to generate a prediction, $f_{\bm \theta^*}(\bm x^*)$, for the true response, $y^* \sim p(y | \bm x^*)$. The uncertainty in our prediction can be divided into two components: aleatoric and epistemic uncertainty. 

\textbf{Epistemic uncertainty} arises from a lack of data sufficiently similar to $\bm x^*$, resulting in a lack of knowledge necessary for $f$ to approximate $p(y | \bm x)$ at $\bm x^*$. Epistemic uncertainty can be reduced by gathering more data near $\bm x^*$, so it is often referred to as \textit{model uncertainty}. 

\textbf{Aleatoric uncertainty} refers to irreducible error, capturing the inherent ambiguity in the relationship between $\bm{x}$ and $y$. For example, we may know a scale is accurate up to 0.5 g after calibrating against a reference material, but even with that information, a new measurement can only estimate the true weight to within 0.5 g. While collecting more samples helps quantify measurement error, the accuracy of any single measurement is limited by the scale itself.

\textbf{Scoring rules} are the gold standard for assessing probabilistic forecasts. Consider a sample space $\Omega$ and family of probabilistic forecasts $\mathcal{P}$. A scoring rule, $S$, is a function $S:\Omega\times\mathcal{P} \rightarrow \overline{\mathbb {R}}$, where $\overline{\mathbb {R}}=[-\infty,\infty]$, that measures the quality of predictions made by $P\in \mathcal{P}$ when data was generated from a distribution $Q\in\mathcal{P}$. Letting $S(P,Q) = \mathbb{E}_{y\sim Q}[S(P,y)]$, scoring rules such that $S(P,Q) \geq S(Q,Q)$ with equality holding iff $P = Q$ are said to be strictly proper \citep{gneiting2007scoringrules, gneiting2014probforecasts}. Strictly proper scoring rules are the gold standard for comparing probabilistic forecasts and include the common negative log-likelihood and Brier score. To make practical use of scoring rules, we use sampled-based estimates, $\frac{1}{n}\sum_{i=1}^n S(P,y_i)$. 

\textbf{UQ for neural networks} traditionally relies on neural network ensembles \citep{lakshminarayanan2017deepensembles} or Bayesian methods that sample network weights with techniques like Markov Chain Monte Carlo (MCMC) \citep{neal2012bayesiandl}, Monte Carlo dropout \citep{gal2016dropout} or approximate the posterior of network weights \citep{chen2014sghmc, ritter2018kfac, maddox2019swag}. These methods involve either training the network multiple times or performing several passes through a single network, both of which are computationally intensive. 

\textbf{Deterministic UQ for neural networks} refers to methods that estimate uncertainty using a single pass through a network by leveraging
outputs from the network’s feature extractor \citep{amersfoot2020DUQ,liu2020sngp,mukhoti2023ddu}. These methods rely on the feature extractor being both \textit{sensitive} and \textit{smooth}, formalized as a bi-Lipschitz constraint \citep{amersfoot2020DUQ, liu2020sngp, mukhoti2023ddu},
\begin{align}\label{eqn:bilipschitz} 
L_1  d_{\mathcal{X}}(\bm x_1, \bm x_2) \leq \lVert h(\bm x_1)-h(\bm x_2)\rVert_{H}\leq L_2 d_{\mathcal{X}}(\bm x_1, \bm x_2), \end{align}
where $L_1>0$ and $L_2>0$ represent sensitivity and smoothness, respectively. Sensitivity ensures meaningful changes in the input space are reflected in feature space, while smoothness promotes generalization.

In residual layers, $h_{l}(\bm x) = a(\bm A_l \bm x + \bm b_l) + \bm x$, \cref{eqn:bilipschitz} can be related to a bound on the spectral norm of the weight matrices, $\sigma(\bm{A}_l)$ \citep{liu2020sngp, bartlett2018smooth}. This relationship suggests that modifying training to incorporate \textit{spectral normalization} (SN) \citep{miyato2018SN} may enable the feature extractor to satisfy the constraint in \cref{eqn:bilipschitz}. Incorporating SN into training has become a standard approach for deterministic UQ methods \citep{mukhoti2023ddu, liu2020sngp}.

Without modifying training to use SN, the network risks feature collapse, where distinct inputs map to identical feature maps, degrading uncertainty quantification (UQ) by reducing the network's ability to distinguish between similar and different data points.

An alternative solution to feature collapse in deterministic UQ involves ensembling intermediate layers \citep{jimenez2023dve}. The approach assumes that, while the constraint in \cref{eqn:bilipschitz} may be violated, it could hold when using the output of an earlier layer. However, this method has only been demonstrated for regression tasks.

Assuming a well structured feature space, we also need a distance-aware output layer. The Spectral-normalized
Neural Gaussian Process (SNGP) \citep{liu2020sngp} uses a Gaussian process as the output layer, while deep deterministic uncertainty (DDU) \citep{mukhoti2023ddu} uses Gaussian discriminate analysis for quantifying epistemic uncertainty and uses the network's predictive entropy for quantifying aleatoric uncertainty. Deterministic uncertainty quantification (DUQ) \citep{amersfoot2020DUQ} uses RBF networks. 

\subsection{Feature geometry}\label{sec:feature_geometry}
We now discuss the geometry of feature space, $\mathcal{H}$, from two perspectives: neural collapse and feature collapse.

\textbf{Neural Collapse} \citep{papyan2020neuralcollapse} is a phenomenon observed during the \textit{terminal phase of training} (TPT), when training error has saturated but training loss continues to decrease, and is characterized by a consistent structure in both the linear classifier, $g$, and the feature space, $\mathcal{H}$. Neural collapse is characterized by the following criteria:
\begin{itemize}
    \item $\mathcal{NC}1$: Vanishing within-class variance of activations.
    \item $\mathcal{NC}2$: Centered classwise activation means have equal length, equal angle between them and are pairwise maximally spread.
    \item $\mathcal{NC}3$: Up to scaling, class-wise activation means and the weights of $g$ converge. 
    \item $\mathcal{NC}4$: $g$ converges to a nearest-class-center classifier. 
\end{itemize}
While neural collapse was initially observed at the penultimate layer, intermediate neural collapse has been shown to occur \citep{rangamani2023intfeatcol} as well.

\textbf{Feature Collapse} describes unique inputs being mapped to identical feature maps. The \textit{smoothness} of a neural network describes the degree of continuity in the mapping from input to feature space, while the \textit{sensitivity} of the network describes how the network's features respond to perturbations in the inputs. A network that is both smooth and sensitive will satisfy a bi-Lipschitz constraint as in \cref{eqn:bilipschitz}, but a network with feature collapse will not satisfy this constraint. 

\textbf{Feature geometry and deterministic UQ} are closely related, as the bi-Lipschitz requirement imposes constraints on the geometry of the feature space. However, little work has been done to leverage results from neural collapse to assess feature collapse in a network, and thus evaluate the effectiveness of deterministic UQ methods without relying on SN. 

For example, relevant to us, $\mathcal{NC}1$ characterizes the network's propensity to map inputs from the same class to similar, if not identical, feature maps. While $\mathcal{NC}4$ characterizes the separability of different classes at a particular layer. Together these two metrics share a seemingly similar role to the sensitivity and smoothness of a network. In \cref{sec:psc} we use this perspective to directly relate feature collapse and neural collapse. 

\section{Related work}
We now contextualize our work by contrasting it with deterministic deep UQ methods that directly address feature collapse. In \cref{tab:methods_compare}, we compare our work to existing methods to highlight that our approach is the first classification-compatible method to allow for deep deterministic UQ without modifying training. 

Additionally, the methods in \cref{tab:methods_compare} that work for classification are limited to particular network structures. Both SNGP and DDU are limited to networks with residual layers and DUQ was built with RBF networks \citep{lecun1998gradient} in mind. However, our approach does not require residual layers to be present and, as shown in \cref{sec:experiments}, we are able to augment VGG16 to perform as well as Resnet18 trained with SN. Our work demonstrates that deterministic UQ methods apply to a broader range of networks than previously suggested.

While the deep Vecchia ensemble (DVE) \citep{jimenez2023dve} also works for networks without residual layers, our work applies to classification while DVE is only relevant to regression. Additionally, our approach of selecting layers using measures of neural collapse is more streamlined than DVE, which ensembles models built on all the intermediate layers.

\begin{table}
    
    \centering
    \begin{tabular}{cccc}
        \toprule
        Method & F.C. Soln.& Retrain & Task \\
        \cmidrule(lr){1-1}\cmidrule(lr){2-2}\cmidrule(lr){3-3}\cmidrule(lr){4-4}
        SNGP \citep{liu2020sngp}, DDU \citep{mukhoti2023ddu} & SN & Yes & Cl. \\
        DUQ \citep{amersfoot2020DUQ} & Jacob. & Yes & Cl. \\
        DVE \citep{jimenez2023dve} & Skip & No & Reg. \\
        Ours & Skip & No & Cl. \\
        \bottomrule
    \end{tabular}
    \caption{\textbf{Our method enables deterministic UQ for classification without the need for retraining.} In this table, ``F.C. Soln.'' refers to the method used to address feature collapse (e.g., spectral normalization, SN), and ``Task'' indicates whether the model is applied to classification (Cl.) or regression (Reg.).}\label{tab:methods_compare}
\end{table}

\section{Probabilistic skip connections (PSCs)}\label{sec:psc}

\begin{figure*}
    \centering
    \includegraphics[width=1.0\linewidth]{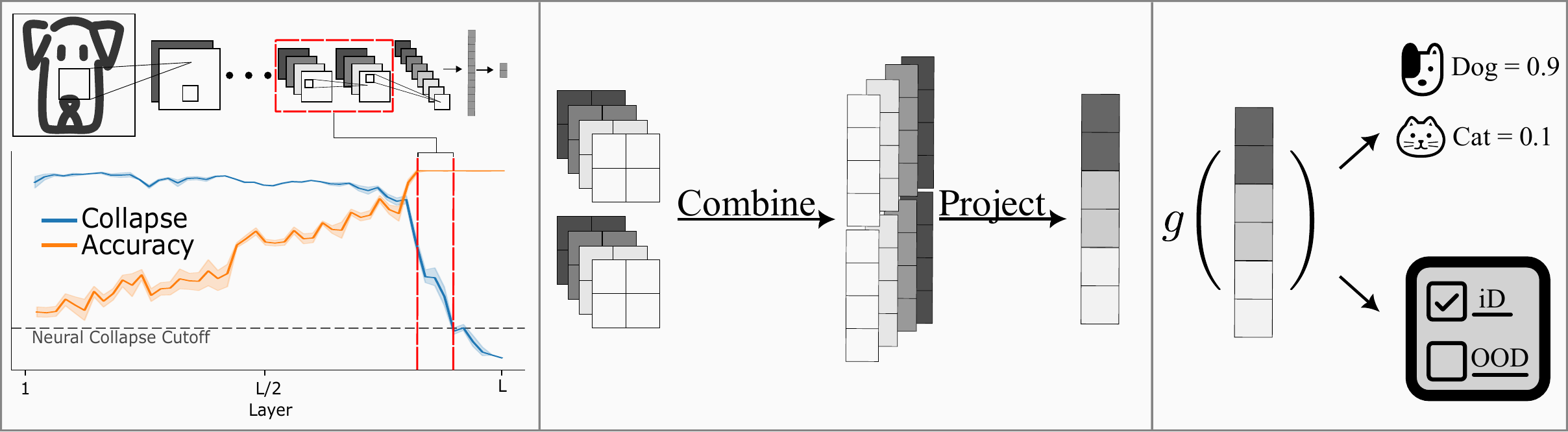}
    \caption{\textbf{Adding PSCs to a pretrained model involves measuring collapse (left), projecting intermediate layers to a feature vector (center), and fitting a probabilistic model to the feature vector (right).} The first step is to find a subset of the layers that tradeoff accuracy and collapse, higher is better for both accuracy and collapse. Once we have those layers we then combine them and project to a single feature vector. Finally, we fit a probabilistic model to that feature vector which can identify iD vs OOD and make class predictions.}
    \label{fig:workflow}
\end{figure*}

The constraint in \cref{eqn:bilipschitz} is typically enforced by modifying network training; however, we propose instead to identify an intermediate layer $h^{(j)}$ for some $j \in \{1, \dots, L\}$ that already meets the bi-Lipschitz requirement in \cref{eqn:bilipschitz}.

Assuming such a subnetwork exists, we define a new model $f'(\bm x) := g' \circ t \circ h^{(j)}(\bm x)$, where $t$ maps the intermediate representations of $h^{(j)}$ to a lower-dimensional feature space and $g'$ is a new output layer. Notice that $g'$ can be chosen from any existing deterministic UQ methods, such as a Gaussian process (as in SNGP), a Gaussian mixture model or linear model (as in DDU), or other distance-aware models as defined in \citet{liu2020sngp}.

Our workflow for finding the layer $j$, getting the projection function $t$ and finally using the new distance aware $g'$ is illustrated in \cref{fig:workflow}. The left panel of \cref{fig:workflow} shows the identification of the layer using \textit{neural collapse}, which is discussed in \cref{sec:choosing_layer}. The center panel represents the function $t$, as discussed in \cref{sec:processing_intermediates}. Finally, the right  panel shows the function $g'$ (discussed in \cref{sec:linear_layer_psc}) being used to determine that a sample is iD and then subsequently making a prediction for that input. 

\subsection{Choosing layer: collapse-accuracy trade-off}\label{sec:choosing_layer}
The leftmost panel of \cref{fig:workflow} shows accuracy and collapse changing as function of layer depth. Below we make it clear what exactly is being computed in \cref{fig:workflow}, discuss how to use collapse and accuracy to choose the location of the PSC and justify our use of collapse and accuracy metrics as proxies for sensitivity and smoothness.

\textbf{Collapse} at layer $l$ is quantified by considering an approximation of $\mathcal{NC}1$ from \citet{rangamani2023intfeatcol} and is denoted $\mathcal{NC}1$ for the remainder of the paper. To begin, for layer $l$ and class $c$ we compute the class-wise feature means and the grand feature mean, 
\begin{align*}
    \bm \mu^{l}_{c}=\frac{1}{N}\sum_{i\in \mathcal{I}_c}^N h^{(l)}(\bm x_i), \hspace{0.5cm}
    \bm \mu^{l}_{G}&=\frac{1}{C}\sum_{c=1}^{C}\bm \mu^{l}_{c}
\end{align*}
where $\mathcal{I}_c\subset \mathcal{I}=\{1,...,N\}$, denote the instances in class $c$. We then compute the within class and total covariance about these means, 
\begin{align*}
\Sigma^{l}_{W}&=\frac{1}{NC}\sum_{c=1}^{C}\sum_{i\in\mathcal{I}_{c}}^{N} (h^{(l)}(\bm x_{i})-\bm \mu_{c}^{l})(h^{(l)}(\bm x_{i})-\bm \mu_{c}^{l})^{T}\\
\Sigma^{l}_{T}&=\frac{1}{NC}\sum_{c=1}^{C}\sum_{i\in\mathcal{I}_{c}}^{N} (h^{(l)}(\bm x_{i})-\bm \mu_{G}^{l})(h^{(l)}(\bm x_{i})-\bm \mu_{G}^{l})^{T}.
\end{align*}

Finally, we can use the within and total covariance to compute the approximation $\mathcal{NC}1\approx Tr(\Sigma_{W})/Tr(\Sigma_{T})$. A layer exhibits \textit{feature variability suppression} if $Tr(\Sigma_{W})/Tr(\Sigma_{T}) < \epsilon=0.2$. We use feature variability suppression as a proxy for feature collapse and we refer to $\epsilon$ as the ``Neural Collapse Cutoff" for simplicity. 

\textbf{Accuracy} of layer $l$ is quantified by measuring the accuracy of a nearest centroid classifier (NCC) that uses $h_{(l)}(\bm x)$ to classify observations and is denoted by $\mathcal{NC}4$ for the remainder of the paper. In this work we use the NCC accuracy of the validation set. 

\textbf{Trading off collapse and accuracy} is done by finding the layer with the highest possible NCC accuracy such that $\mathcal{NC}1>\epsilon$, taking both the layer before and after collapse if $\mathcal{NC}1$ is especially close to $\epsilon$. We refer to the chosen layer(s) as the \textit{candidate layer(s)}.

In \cref{fig:workflow} the candidate layers have a dashed red box around them in the leftmost panel. In \cref{sec:feature_geometry_experiment}, we carry out experiments to justify using $\mathcal{NC}$1 and $\mathcal{NC}4$ to select the candidate layer.

\textbf{Why use $\mathcal{NC}1$ and $\mathcal{NC}4$?}
Sensitivity ensures that distances in the hidden space correspond to meaningful distances on the data manifold. $\mathcal{NC}1$, which measures the within-class variance relative to the total variance, captures this by indicating whether class-wise representations retain variability ($\mathcal{NC}1 > 0$) or collapse to single points. As $\mathcal{NC}1$ decreases deeper in the network, it reflects increasing sensitivity to class-distinguishing perturbations over within-class ones, akin to having a lower Lipschitz bound at that layer.

Smoothness, on the other hand, ensures that semantically similar inputs are mapped to nearby points. $\mathcal{NC}4$, the accuracy of a nearest-centroid classifier, reflects this by measuring how well feature mappings preserve semantic similarity. High $\mathcal{NC}4$ indicates that the network is primarily sensitive to class-relevant perturbations, resembling an upper Lipschitz bound on the outputs.

\subsection{Processing intermediate representations}\label{sec:processing_intermediates}
Once we have found the candidate layers, we need to combine them if there are multiple, and project to a lower dimensional shape if they are high-dimensional. 

\textbf{Combining layers} involves reshaping them and concatenating. For a convolutional layers $h^{(j)}(\bm x) \in \mathbb{R}^{C\times h\times w}$ and $h^{(j+1)}(\bm x) \in \mathbb{R}^{C\times h'\times w'}$ we would reshape the layers to $\mathbb{R}^{C\times h d}$ and $\mathbb{R}^{C\times h' w'}$, respectively. We then concatenate channel wise to get a combined input of shape $\mathbb{R}^{C\times (h' w'+h w)}$, which we will refer to as the candidate layer as the layers have now been combined. For simplicity, we will let $\bm X$ denote candidate layer and assume $\bm X \in \mathbb{R}^{C\times h d}$.

\textbf{Projecting} the layers is done by first computing the channelwise covariance matrices for the candidate layer, which results in a tensor of covariances $\bm \Sigma_c\in\mathbb{R}^{C\times hw \times hw}$. We then perform a Tucker decomposition \citep{tucker1963decomp}; that is, we approximate the covariance using,
\begin{align*}
    \bm \Sigma_{c} &\approx \mathcal{G}\times_1\bm A \times_2 \bm B \times_3 \bm C,
\end{align*}
where $\times_i$ denotes the $i$-mode product. The matrices $\bm A\in \mathbb{R}^{C\times c_{proj}}$, $\bm B\in \mathbb{R}^{hw\times d_{proj}}$ and $\bm C\in \mathbb{R}^{hw\times d_{proj}}$ are the factor matrices and can be thought of as the principal components in each mode. The tensor $\mathcal{G}\in \mathbb{R}^{c_{proj}\times d_{proj} \times d_{proj}}$ is called the core tensor and its entries show the level of interaction between the different components. Given the Tucker decomposition of $\bm \Sigma_c$, we can then project $\bm X$ onto the factor matrices to get a low-dimensional $\tilde{\bm X}$,
\begin{align*}
    \tilde{\bm X} &= \bm X \times_1\bm A_{:, 0:c_{proj}}\times_2 \bm B_{:, 0:d_{proj}},
\end{align*}
where $\tilde{\bm X} \in \mathbb{R}^{c_{proj} \times d_{proj}}$, and $\bm K_{:, 0:d}$ denotes the first $d$ columns of the matrix $\bm K$. Finally, we can flatten this projection to get the final feature vector $\bm Z \in \mathbb{R}^{c_{proj} d_{proj}}$. 

The primary benefit of the Tucker decomposition is it allows us to choose how much we reduce the channel dimension versus how much we reduce the spatial dimension. This means we must choose two parameters at this stage: the projection dimension of the channels ($c_{proj})$ and the dimension of the spatial projections ($d_{proj}$). In \cref{sec:processing_intermediates_experiment}, we demonstrate the low-dimensional nature of intermediate representations and provide guidance on setting these parameters based on an empirical study. In short, we recommend setting $c_{proj}$ and $d_{proj}$ to be the smallest values such that neither $\mathcal{NC}1$ and $\mathcal{NC}4$ change dramatically.

\subsection{Linear layers as PSCs}\label{sec:linear_layer_psc}
After processing the outputs of an intermediate layer, we now have input-output pairs, $\{\bm Z_i, y_i\}_{i=1}^N$ with $\bm Z_i \in \mathbb{R}^{c_{proj} d_{proj}}$ being the output of the ``Project'' step in \cref{fig:workflow} and $y_i \in \{1,...,C\}$ being the class label. At this point any distance-aware model can be used to make predictions. 

In this work, we decompose $g$ into two components, similar to the approach in DDU \citep{mukhoti2023ddu}. For iD/OOD classification, we use quadratic discriminant analysis, and for iD UQ, we apply a linear model with a Laplace approximation for the posterior of the coefficients. To ensure scalability, we employ the Kronecker-factored Laplace approximation (KFAC) \citep{ritter2018kfac} as implemented in the Laplace package in PyTorch \citep{daxburg2021redux}. However, our linear model and feature density could be substituted with any other probabilistic model; our primary contribution is demonstrating how deterministic UQ can be achieved by leveraging intermediate network layers.
\section{Experiments}\label{sec:experiments}
This section presents experiments to show the following:
\begin{itemize}
    \item Training with SN ensures a sensitive and smooth feature extractor, as reflected by the layerwise values of $\mathcal{NC}1$ and $\mathcal{NC}4$. Additionally, the same network trained without SN exhibits an intermediate layer that is both sensitive and smooth (\cref{sec:feature_geometry_experiment}).
    \item Using the appropriate intermediate layer allows us to use deterministic UQ methods without retraining and with networks that do not have residual layers, with the performance matching that of a network retrained with SN (\cref{sec:ddu_case_study}, \cref{sec:iD_UQ}). 
    \item Intermediate feature vectors can be projected to a lower-dimensional space without undermining sensitivity and smoothness (\cref{sec:processing_intermediates_experiment}).
\end{itemize}
For additional experiments, see \cref{appendix:additional_experiments}.

\subsection{Trade-off between sensitivity and smoothness}\label{sec:feature_geometry_experiment}
In this section, we demonstrate that training a network with SN adjusts the values of $\mathcal{NC}1$ and $\mathcal{NC}4$ at the penultimate layer in line with the stated goal of enforcing the bi-Lipschitz condition described in \ref{eqn:bilipschitz}. Furthermore, without SN, we observe that intermediate layers exhibit a trade-off between sensitivity and smoothness, with a subset of layers achieving both properties simultaneously.

\begin{figure}
    \centering
    \includegraphics[width=1.0\linewidth]{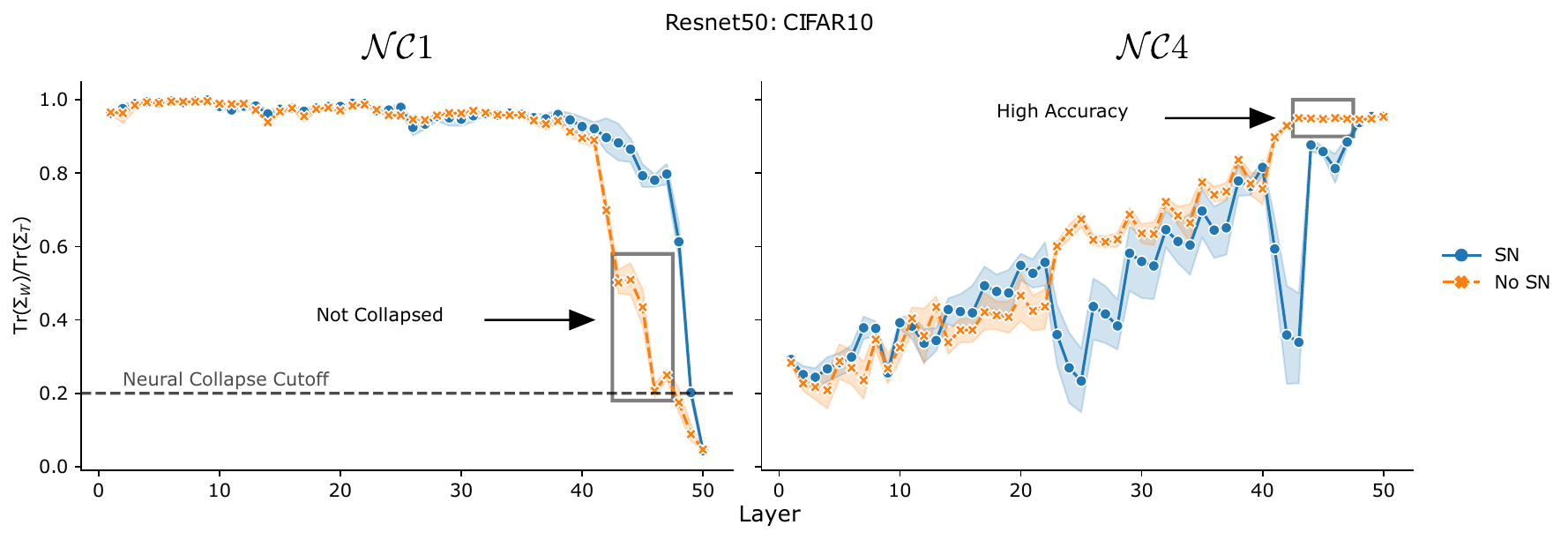}
    \caption{\textbf{SN delays neural collapse to the final layer, whereas without SN, intermediate layers behave similarly to the penultimate layer of a network trained with SN.} (Left) SN pushes the point where $\mathcal{NC}_1$ drops below 0.2 to the final layer. Without SN, there is a region where $\mathcal{NC}_1$ remains above 0.2, but its values differ from those observed with SN. (Right) SN reduces the performance of the nearest-centroid classifier (NCC) across all but the last two layers, while without SN, more layers exhibit NCC accuracy comparable to the final prediction. (Both Panels) Boxes highlight layers that have not yet collapsed but maintain high NCC accuracy, indicating that even without SN, some layers are both \textit{sensitive} and \textit{smooth}.}
    \label{fig:sn_neural_collapse}
\end{figure}

We begin by training a ResNet50 on CIFAR10, both with and without SN. For each model, we compute $\mathcal{NC}1$ and $\mathcal{NC}4$ values at the outputs of every convolutional layer, the final average pooling layer, and the linear classifier, using the implementation from \citet{rangamani2023intfeatcol}. The results are presented in \cref{fig:sn_neural_collapse}, with the left panel showing $\mathcal{NC}1$ values across layers and the right panel showing $\mathcal{NC}4$ values. In \cref{fig:sn_neural_collapse}, the feature extractor’s outputs correspond to layer 49 and the color denotes whether or not we used SN.

The left panel of \cref{fig:sn_neural_collapse} shows that training with SN (blue) delays collapse to the final layer, resulting in a feature extractor that remains sensitive ($\mathcal{NC}1 > \epsilon$). Meanwhile, the right panel indicates that the same feature extractor achieves high nearest-centroid accuracy, confirming that it is both sensitive and smooth. In contrast, training without SN (orange) allows collapse to occur earlier in the network, but the nearest-centroid accuracy of these earlier intermediate layers matches that of the final layer. Notably, a subset of intermediate layers in the network without SN exhibits both high sensitivity ($\mathcal{NC}1 > \epsilon$) and high nearest-centroid accuracy. This shows that, by selecting different layers, sensitivity and smoothness can be traded off as needed. Our candidate layer is chosen from this subset, meaning our approach can be viewed as truncating the network.

The performance of the nearest-centroid classifier (NCC) at intermediate layers has been noted in prior work \citep{benshaul2022nearest, galanti2022implicit, rangamani2023intfeatcol}, with \citet{galanti2022implicit} suggesting that this behavior indicates the network's ``effective depth'' is lower than its nominal depth. From this perspective, SN can be seen as increasing the effective depth of the network, which may provide additional benefits beyond uncertainty quantification, such as improved generalization \citep{galanti2022implicit} or enhanced supervised pretraining.

\subsection{DDU case study}\label{sec:ddu_case_study}
This experiment demonstrates that DDU \citep{mukhoti2023ddu} remains effective when SN is replaced by an intermediate layer from a network trained without SN. Furthermore, we show that intermediate layers can be effective for UQ even in the absence of residual connections, which are typically required when using the penultimate layer for deterministic UQ.

Our experiment builds on \citet{mukhoti2023ddu}, utilizing Dirty-MNIST (a combination of MNIST and Ambiguous-MNIST) as the in-distribution dataset and FashionMNIST as the out-of-distribution (OOD) dataset. Notably, Ambiguous-MNIST contains samples with higher aleatoric uncertainty compared to MNIST, while FashionMNIST (FMNIST) examples exhibit high epistemic uncertainty.

\begin{figure}[ht]
    \centering
    \includegraphics[width=1.0\linewidth]{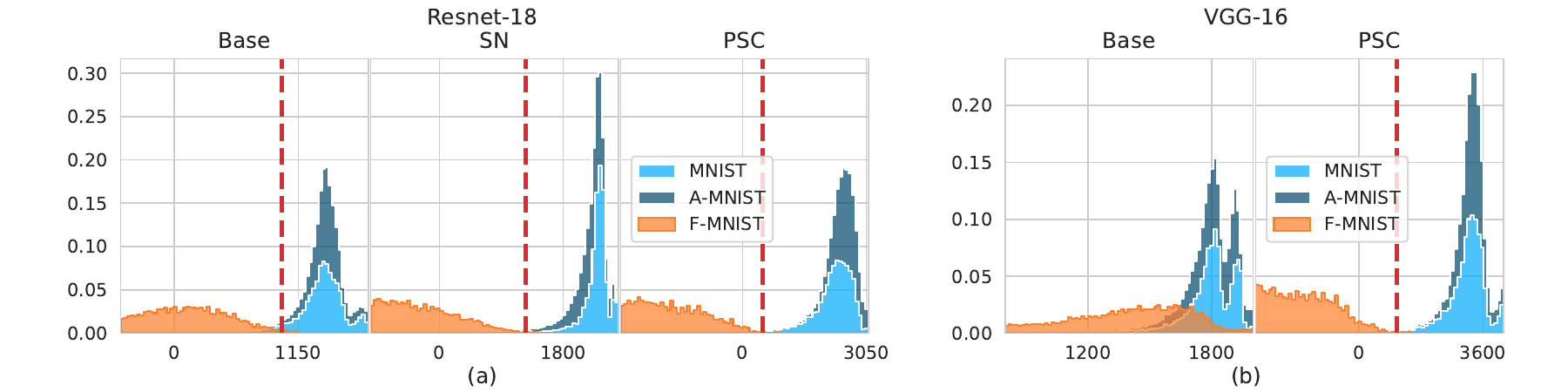}
    \caption{(a) \textbf{PSC feature density separates iD and OOD just as well as when using SN.} The panels each show the probability of the feature vector under GDA trained on the embeddings of the training data, but each panel differs in how the feature vectors are computed. The base network's original feature density uses no SN (left) and can be improved equally well by using SN (middle) or  an intermediate layer (right). (b) \textbf{Intermediate feature density also works for networks without residual connections and no SN.} The feature density of VGG-16 (left) conflates iD and OOD but using an intermediate layer (right) separates iD and OOD.}
    \label{fig:dirty_mnist_density}
\end{figure}

We train two networks on Dirty-MNIST, one with SN and one without. For both networks, we fit classwise multivariate normal distributions to the feature vectors $h(\bm{x})$, enabling us to evaluate the \textit{feature density} $p(h(\bm{x}))$ for test instances. These densities, labeled "Base" (without SN) and "SN" (with SN), are shown in \cref{fig:dirty_mnist_density}. For the network without SN, we also compute an intermediate feature vector by following the first two panels of \cref{fig:workflow} and then adding a final PCA-based projection. This method is referred to as "PSC" in \cref{fig:dirty_mnist_entropy}. Additionally, we fit a multinomial logistic regression model to the intermediate layer which we use to replace the original network predictions. Feature densities are used to identify out-of-distribution (OOD) instances, while in-distribution (iD) predictions use network outputs. Full implementation details and results are in \cref{appendix:ddu_details} and \cref{appendix:dirty_mnist}. Below are the main results of this experiment.

\textbf{We can distinguish epistemic and aleatoric uncertainty using intermediate layers}: \cref{fig:dirty_mnist_density} panel (a) shows the feature densities of the different dataset for a Resnet-18 fit to Dirty-MNIST. The left panel corresponds to using the original network, the center panel corresponds to the network trained with SN, and the right panel corresponds to the original network augmented with a PSC. We see that the feature density when using the intermediate layer distinguishes between iD and OOD just as well as when using SN, with both providing a modest improvement over the original network. In \cref{appendix:dirty_mnist} we demonstrate that predictive entropy of the PSC has similar properties to when using SN.

In \cref{tab:dirty_mnist_auroc}, we present the AUROC values when using feature space density to compare iD and OOD examples, and we also present the AUROC values when using predictive entropy to compare iD-ambiguous to iD-clean. For both tasks, we see the AUROC of PSC matching the AUROC found when using SN or exceeding it, which confirms that using the intermediate space is a good substitute for SN.

Together, \cref{fig:dirty_mnist_density} and \cref{tab:dirty_mnist_auroc} demonstrate that feature density captures epistemic uncertainty but not aleatoric uncertainty, whereas predictive entropy captures aleatoric uncertainty. This behavior, observed in \citet{mukhoti2023ddu}, is replicated here; however, we achieve it solely using intermediate layers without relying on SN.

\begin{table}
    \centering
    \begin{tabular}{lll}
    \toprule
    &\multicolumn{2}{c}{AUROC ($\uparrow$)}\\
    \cmidrule(lr){2-3}
    Model & iD/OOD & iD/iD-Ambig.\\
    \cmidrule(lr){1-1}\cmidrule(lr){2-2}\cmidrule(lr){3-3}
    Resnet-18 & 0.99 & 0.97\\
    Resnet-18+SN & 0.99 & 0.84\\
    Resnet-18+PSC & 0.99 & 0.93\\
    \cmidrule(lr){1-3}
    VGG-16 & 0.93 & 0.98\\
    VGG-16+SN & N/A & N/A\\
    VGG-16+PSC & 1.0 & 0.95\\
    \bottomrule
    \end{tabular}
    \caption{\textbf{OOD scores when using PSCs match those when using SN and PSCs can be applied to networks without residual layers.} The AUROC for Dirty-MNIST vs. FashionMNIST (iD/OOD) and MNIST vs. Ambiguous-MNIST (iD/iD-Ambig.) for different models. Models below solid line do not have residual connections, so classic DDU is not applicable.}
    \label{tab:dirty_mnist_auroc}
\end{table}

\textbf{DDU works without residual connections}: In \cref{fig:dirty_mnist_density} panel (b) we replace Resnet-18 with VGG-16 \citep{simonyan14vgg}, which has no residual connections, and rerun the experiment. We see that the original network has a poorly structured feature space as the feature density for iD and OOD instances overlap, but the intermediate feature space is surprisingly well behaved and appears as well separated as Resnet-18 trained with SN. This is confirmed by examining the final rows of \cref{tab:dirty_mnist_auroc}, where the AUROC of VGG-16+PSC matches the best models in separating iD and OOD. Likewise, the predictive entropy of the PSC can be effectively used to when separating Ambiguous-MNIST and MNIST. Again, there are no residual connections in this model, yet DDU performs well. 

\subsection{Intermediate representations have exploitable low-rank structure}\label{sec:processing_intermediates_experiment}
This experiment demonstrates that projecting intermediate representations to a lower-dimensional space preserves the \textit{smoothness} and \textit{sensitivity} properties identified at the \textit{candidate layer} in \cref{sec:feature_geometry_experiment}. Consequently, these projected feature vectors can replace the typically high-dimensional intermediate feature vectors without causing feature collapse or compromising predictive accuracy.

\textbf{Projecting preserves feature geometry:} We begin by training various networks on different datasets and identifying the candidate layer for a PSC using our two neural collapse metrics. Next, we project the intermediate representations of these candidate layers to a lower-dimensional space and recompute the neural collapse metrics on the projected layers.

\cref{tab:intermediate_projections} presents the values of $\mathcal{NC}1$ and $\mathcal{NC}4$ computed before and after projection. Across all combinations, the results show that both before and after projection, there is no collapse ($\mathcal{NC}1 > \epsilon$), and the nearest-centroid accuracy remains nearly unchanged. This indicates that with an appropriate projection, feature collapse can be avoided while maintaining strong predictive performance.

\begin{table}
    \centering
    \begin{tabular}{llllll}
    \toprule
     & &\multicolumn{2}{c}{$\mathcal{NC}1$} & \multicolumn{2}{c}{$\mathcal{NC}4$}\\
     \cmidrule(lr){3-4}\cmidrule(lr){5-6}
    Model&Dataset&Orig.&Proj.&Orig.&Proj.\\
    \cmidrule(lr){1-1}\cmidrule(lr){2-2}\cmidrule(lr){3-3}\cmidrule(lr){4-4}\cmidrule(lr){5-5}\cmidrule(lr){6-6}
    Resnet-18&MNIST&0.28&0.28& 0.99&0.99\\
    WRN&CIFAR10&0.73& 0.75& 0.95&0.95\\
    Resnet50&CIFAR10&0.28& 0.50 & 0.95& 0.94\\
    VGG-16&MNIST&0.26&0.36& 0.97&0.97\\
    \bottomrule
    \end{tabular}
    \caption{\textbf{Projection of intermediate representations to a lower-dimensional space preserves sensitivity ($\mathcal{NC}1$) and smoothness ($\mathcal{NC}4$).} Neural collapse metrics before (Orig.) and after projection (Proj.) for various networks and datasets.}
    \label{tab:intermediate_projections}
\end{table}

\textbf{Projection results are robust to hyperparameters:} The results in \cref{tab:intermediate_projections} depend on the projection parameters $c_{proj}$ and $d_{proj}$, as described in \cref{sec:processing_intermediates}. To assess robustness, we repeat the previous experiment while varying the projection dimensions to evaluate the impact of these parameters on the results.

\begin{figure}
    \centering
    \includegraphics[width=1.0\linewidth]{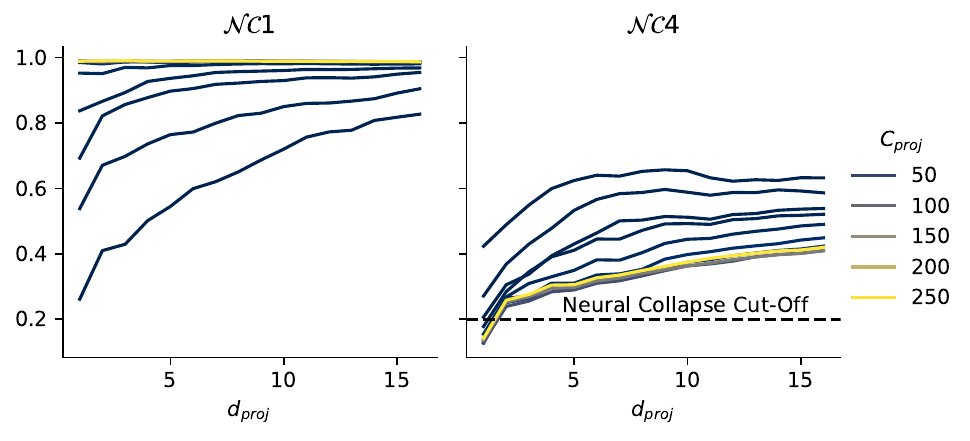}
    \caption{\textbf{For a wide range of $c_{proj}$ and $d_{proj}$, neural collapse is avoided, and NCC accuracy remains high, demonstrating the robustness of the projection method.} Neural collapse metrics for ResNet-18 on MNIST are shown, with the left panel displaying $\mathcal{NC}1$ values and the right panel showing $\mathcal{NC}4$ values. In both panels, the x-axis represents $d_{proj}$, and color indicates $c_{proj}$.}
    \label{fig:c_proj_d_proj_nc_metrics}
\end{figure}

\cref{fig:c_proj_d_proj_nc_metrics} shows the impact of varying $c_{proj}$ and $d_{proj}$ on the neural collapse metrics for ResNet-18 on MNIST, with similar trends observed across other networks and datasets. As expected, there are some values of $c_{proj}$ and $d_{proj}$ where a previously uncollapsed layer ($\mathcal{NC}1 > \epsilon$) becomes collapsed ($\mathcal{NC}1 < \epsilon$). However, for a broad range of parameter values, no collapse is observed, and the nearest-centroid classifier (NCC) accuracy remains consistently high.

\subsection{iD UQ performance for PSCs matches using SN}\label{sec:iD_UQ}
This experiment demonstrates that adding PSCs to a network without SN can achieve in-distribution (iD) UQ performance comparable to a network trained with SN. To evaluate this, we compare the iD performance of models under different configurations.

First, we train a ResNet28x10 on the CIFAR-10 dataset, both with and without SN. Using these networks, we make predictions on the test set and evaluate their performance. Next, we augment the network without SN with a PSC following the method described in \cref{sec:linear_layer_psc}. The augmented network is then used to make predictions on the same test set. For all three models—base network without SN, network with SN, and network without SN but with PSC—we evaluate accuracy, negative log-likelihood (NLL), and expected calibration error (ECE). This process is repeated across 10 different train-validation splits.

The results of this experiment, presented in \cref{tab:cifar10}, show the mean and standard error across the ten different seeds. While the original network achieves the highest accuracy, PSCs match or outperform using SN in all UQ metrics, with a particularly significant improvement in expected calibration error (ECE). Additional results can be found in \cref{appendix:additional_experiments}.

\begin{table}
    \centering
    \resizebox{\linewidth}{!}{%
    \begin{tabular}{ccccc}
    \toprule

    Method & 
    Requires Retraining & Accuracy ($\uparrow$) & NLL ($\downarrow$) & ECE ($\downarrow$)\\ \midrule 
    
    DNN & N & \textbf{96.00} $\pm$ \textbf{0.01} & 0.15 $\pm$ 0.001 & 0.023 $\pm$ 0.001 \\

    DNN-PSC (Ours)& N   & 95.80 $\pm$ 0.01 & \textbf{0.136} $\pm$ \textbf{0.001} & \textbf{0.005} $\pm$ \textbf{0.000} \\
    
    DNN-SN &Y  & 96.00 $\pm$ 0.03 & 0.153 $\pm$ 0.002 & 0.022 $\pm$ 0.001 \\
    
    \bottomrule
    \end{tabular}
    }
    \caption{\textbf{PSC achieves in-distribution (iD) UQ performance comparable to using SN for Wide-ResNet-28-10, without requiring retraining.} The table reports accuracy, negative log-likelihood (NLL), and expected calibration error (ECE) for three methods: the base DNN without SN, the DNN-PSC (our method), and the DNN with SN. Results are presented as the mean and standard error over ten random seeds.}
    \label{tab:cifar10}

\end{table}

\section{Conclusion}
In this paper, we introduced probabilistic skip connections (PSCs) for deterministic uncertainty quantification (UQ) in classification networks without retraining with spectral normalization. By leveraging neural collapse metrics, we located an intermediate layer that preserves feature sensitivity and smoothness, placing a PSC there to mitigate feature collapse. Our method achieves UQ performance comparable to networks retrained with spectral normalization while extending deterministic UQ to a broader range of architectures.

We validated our approach through comprehensive experiments, showing that spectral normalization effectively increases the network’s depth, whereas our method leverages the network’s initial effective depth. We also demonstrated that existing deterministic methods, when applied to our truncated network, match the OOD and UQ performance of networks retrained with spectral normalization. Moreover, we confirmed the robustness of PSCs under various projection parameters.

Our approach enables high-quality UQ without retraining, offering a scalable solution. However, there are limitations. Standardizing projection dimensions for common architectures would facilitate a true drop-in replacement for existing models, and further exploration of metrics for feature collapse could optimize projection sizes and reduce parameter sweeping.

Future work could extend PSCs to other UQ tasks, such as reinforcement learning or time-series forecasting. Refining techniques for identifying intermediate layers that maintain feature sensitivity and smoothness could further enhance the robustness and versatility of our approach across diverse architectures and domains.

{
    \small
    \bibliographystyle{plainnat}
    \setlength{\bibsep}{.1pt}
    \bibliography{ref}
}

\appendix
\clearpage
\setcounter{page}{1}

\twocolumn[
    \centering
    \Large
    \vspace{0.5em}Supplementary Material \\
    \vspace{1.0em}
]

\startcontents[appendix]
\printcontents[appendix]{l}{1}{\textbf{Supplementary Material Contents:}}
\section{Method details}
This section provides details on our proposed methodology. The pseudo-code in \cref {fig:psc_pseudo_code} summarizes the steps necessary to augment a pretrained neural network with a PSC. The code between loading data and forming the prediction is detailed in the sections below.

\begin{figure}[ht]
\centering
\begin{lstlisting}
# load model and data.
net = load_pretrained_net(...)
train, val, test = get_data(...)

# identify candidate layer.
candidate = measure_collapse(
    net, train, val
)

# fit projection.
projection = fit_projection(
    net, 
    train, 
    "tucker"
)

# fit PSC.
net = fit_psc(
    net=net, 
    projection=projection, 
    psc="ddu", # can replace with SNGP
    candidate=candidate, 
    data=train
)

# predict.
y_hat, ood_score = net.predict(test)

\end{lstlisting}
    \caption{\textbf{Pseudo-Code for working with pretrained networks highlights the drop-in nature of PSCs.}}
    \label{fig:psc_pseudo_code}
\end{figure}

\subsection{Measuring collapse}\label{appendix:methods_measuring}
\cref{fig:psc_pseudo_code} highlights that we choose the \textit{candidate layer} using measures of neural collapse. In particular we meausure $\mathcal{NC}1$ and $\mathcal{NC}4$ \textit{after} each layer. This is different than the work of \citet{rangamani2023intfeatcol} where the metrics are measured before each layer. This difference arises due to our different motivations, in \citet{rangamani2023intfeatcol} the focus is on drawing parallels between intermediate feature collapse and feature collapse in the penultimate layer. However, in our work we focus on finding the layer that results in a feature vector with little collapse but high predictive performance. Computing both metrics can be done using simple modifications of the code from \citet{rangamani2023intfeatcol}.

\subsection{Projecting intermediate representations}\label{appendix:methods_processing}
For every call to the model we need to project the intermediate representations to get a feature vector. The steps to get this projection are shown in \cref{alg:reshape_scale_project}. This algorithm assumes we have parameters $\bm \Sigma$, $\bm \mu$, $\bm \Sigma_{\bm Z}$, $\bm \mu_{\bm Z}$ in addition to $\bm A$, $\bm B$. The first set of these parameters correspond to the moments of $\bm X$ and to the moments of $\bm Z$. The parameters $\bm A$ and $\bm B$ correspond to the factor matrices from the Tucker decomposition of the channel-wise covariance of $\bm X$.

All of these parameters must be learned from the data, and this is highlighted in the psuedo-code by the call to the function "fit\_projection" which returns these parameters for future calls to \cref{alg:reshape_scale_project}. Details on computing the moments are given in \cref{appendix:channel_wise_moments}. The details on the subroutines called in \cref{alg:reshape_scale_project} and computing $\bm A$ and $\bm B$ are given below.

\begin{algorithm}
\begin{algorithmic}
\STATE \textbf{Input:} $\bm X, \bm \Sigma, \bm \mu, \bm \Sigma_{\bm Z}, \bm \mu_{\bm Z}, \bm A, \bm B$
\STATE $\bm X \gets \text{reshape}(\bm X)$ \COMMENT{E.g. \cref{fig:workflow} center panel.}
\STATE $\bm X \gets \text{scale}(\bm X, \bm \Sigma, \bm \mu)$
\STATE $\bm Z \gets \text{project}(\bm X, \bm A, \bm B)$
\COMMENT{Tucker.}
\STATE $\bm Z \gets \text{scale}(\bm Z, \bm \Sigma_{\bm Z}, \bm \mu_{\bm Z})$ 
\COMMENT{Optional.}
\STATE \textbf{Output:} $\bm Z$
\end{algorithmic}
\caption{\textbf{Processing intermediate representations involves three steps: reshape, scale and project}.}\label{alg:reshape_scale_project}
\end{algorithm}

\subsubsection{Steps in projection algorithm}
\textbf{Reshape} The first operation involves reshaping the input. For the output of a single convolutional layer we use 1-mode matricization \citep{kolda2009tensordecomp} which simply turns inputs of shape $\mathbb{R}^{C\times h \times w}$ to shape $\mathbb{R}^{C\times h w}$ where we assume row-major flattening of each channel (i.e. $\text{vec}(\bm A^T)$ with $\bm A\in\mathbb{R}^{h\times w}$). If we want to combine multiple convolutional layers then we reshape each of them as describe and then concatenate them channel-wise.

\begin{figure*}
    \centering
    \includegraphics[width=1.0\textwidth]{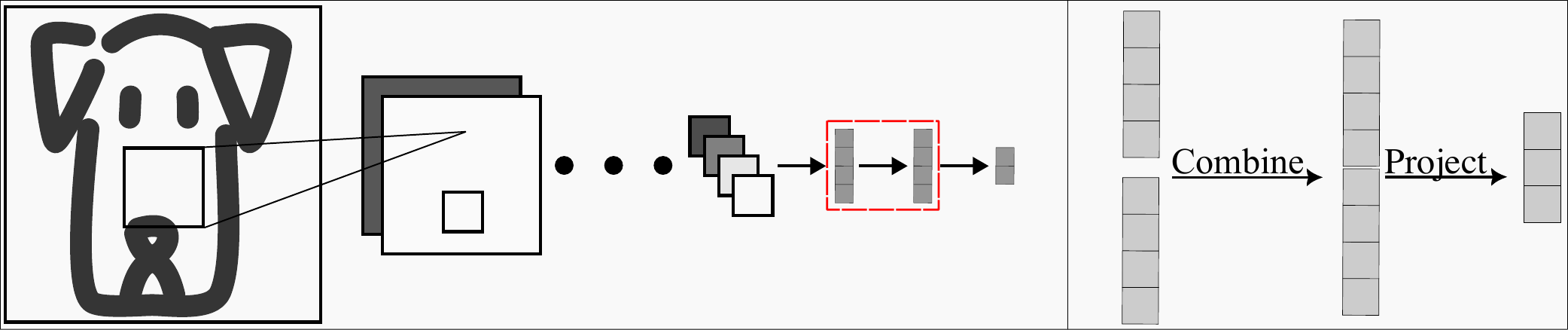}
    \caption{\textbf{We can handle different layer outputs in a similar way to convolutional layers}. The candidate layers of a network are both fully connected layers, and after concatenating and projecting the feature vector can be handled just as before. }
    \label{fig:fc_candidate_layers}
\end{figure*}

\cref{fig:fc_candidate_layers} show that our approach for processing fully connect layers is similar and the resulting feature vector is also quite similar. By processing the inputs this way, the downstream methodology is identical regardless of the type of the candidate layer. 

\textbf{Scale}
The scaling operations assumes inputs of shape $\bm X \in \mathbb{R}^{C \times d}$ with $\bm \mu \in \mathbb{R}^{C \times d}$ and $\bm \Sigma \in \mathbb{R}^{C \times d \times d}$. Each input is then centered and scaled channelwise, 
\begin{align*}
(\bm X_{i,:} - \bm \mu_{c,:}) \text{diag}(\bm \Sigma_{c,:,:})^{-1},
\end{align*}

where $\text{diag}(\cdot)$ returns a diagonal matrix whose values match the diagonal of the input matrix,  $\bm \mu_{c,:}$ represents the channel-wise mean, and $\bm \Sigma_{c,:,:}$ represents the channel-wise covariance. 

\textbf{Project}
When we perform the projection step we use the Tucker decomposition of the covariance matrix output from \cref{alg:moments} but with one exception. Since the output of the intermediate representation has been centered and scaled we first transform the covariance matrices before using the Tucker decomposition. More precisely, for channel $c$ of the covariance tensor we compute the following,
\begin{align*}
    \bm \Sigma_{c,:,:} = \text{diag}(\text{sqrt}(\bm \Sigma_{c,:,:}))^{-1}  \bm \Sigma_{c,:,:} \text{diag}(\text{sqrt}(\bm \Sigma_{c,:,:}))^{-1},
\end{align*}

where $\text{diag}(\text{sqrt}(\bm \Sigma_{c,:,:}))^{-1}$ denotes the diagonal matrix whose elements are the sample standard deviations of each dimension in the $c^{\text{th}}$ channel. Once this transformation has taken place the projection step follows exactly as in \cref{sec:processing_intermediates}.

\subsubsection{Computing channel-wise moments}\label{appendix:channel_wise_moments}
For dataset of $N$ inputs we estimate "channelwise" means and covariances as illustrated in \cref{alg:moments}. In words we use two passes through the data, one to estimate the mean of a hidden layer's output and then another pass to estimate the covariance of the same layer's output. We would like to make two comments on Algorithm \cref{alg:moments}: 

(1) While in practice we vectorize the estimation of the moments (so the loop over channels and the loop over batches would be swapped), Algorithm \cref{alg:moments} is written to emphasize that we do not consider the cross correlation between channels. We ignore cross channel correlation because if we did not, the size of resulting covariance matrix would be too large. For example with Resnet50 on CIFAR10 the covariance matrix of the candidate layer would be of size $131,072 \times 131,072$. By ignoring cross channel covariance the project operation is greatly simplified and fast enough to be of practical use, even with large intermediate representations.

(2) While we use two passes through the training data (for each meand covariance pair), we found that a subset of the full training set could be used to estimate the empirical moments without significant changes in results. However, it was important that the estimate of the mean was as accurate as possible before beginning to estimate the covariance. Using a streaming estimate of the mean along with a streaming estimate of the covariance gave us poor estimates of the feature covariance and this seemed to negatively impact the stability of training PSCs downstream.

\begin{algorithm}[ht]
\caption{\textbf{Channelwise moments are estimated independently in batches}.}\label{alg:moments}
\begin{algorithmic}
\STATE \textbf{Input:} $\bm X \in \mathbb{R}^{N\times C\times d}$

\STATE $\bm \mu \gets \bm{0}, \bm \Sigma \gets \bm{0}$ 

\FOR{$c$ in $\{1,...,C\}$}  

\FOR{$b$ in $B$}
\STATE $\bm \mu_{c,:} \gets \frac{1}{|b|} \sum_{i\in \bm b}\bm X_{i, c, :} - \bm \mu_c$
\ENDFOR

\FOR{$b$ in $B$}  

\STATE $\bm \Sigma_c \gets \bm \Sigma_c + \sum_{i\in \bm b} (\bm X_{i,c,:} - \bm \mu_c)^T(\bm X_{i,c,:} - \bm \mu_c)$
\ENDFOR
\STATE $\bm \Sigma_c \gets \bm \Sigma_c / N$
\ENDFOR

\STATE \textbf{Output:} $\bm \mu, \bm \Sigma$
\end{algorithmic}
\end{algorithm}

\subsection{Using PSCs}\label{appendix:methods_psc}
Once the feature vector has been formed we need a model that can use the feature vector to make predictions. Following \citet{liu2020sngp} we refer to this model as $g$ and we require that $g$ is \textit{distance aware}. For this work we simply use a linear layer that is trained to map the projected intermediate representations, $\bm Z$ in \cref{alg:reshape_scale_project}, to the class label. We then use a Laplace approximation for the posterior of the network weights. Where we further approximatee the covariance of the Laplace approximation using the Kronecker factored Laplace approximation \citep{ritter2018kfac} as implemented in \citet{daxburg2021redux}. For the out-of-distribution prediction we can use quadratic discriminant analysis as documented in \cref{sec:ddu_case_study}.

\section{Experiment details}

\subsection{Network Training}\label{appendix:resnet_training}
For the neural networks in \cref{sec:experiments} we use the same training procedure and networks as in \citep{mukhoti2023ddu}. In short, we use stochastic gradient descent with momentum. The weight decay, momentum and learning rates are set to the values in \cref{tab:resnet_hypers}. During training the learning rate was held constant and then multiplied by a factor of 0.1 at 150 and again at 250 epochs. 

\begin{table}[ht]
    \centering
    \label{tab:resnet_hypers}
    \begin{tabular}{cc}
    \toprule
       Hyperparameter  &  Value \\
       \cmidrule(lr){1-1}\cmidrule(lr){2-2}
       Initial LR  & 0.1\\
       Weight decay & 5e-4\\
       Momentum & 0.9\\
       \cmidrule(lr){1-2}
       Spectral Norm (if used) & 3.0\\
    \bottomrule
    \end{tabular}
    \caption{\textbf{Standard hyperparameters were used during training} and spectral norm was set to 3.0 if it was used.}
\end{table}

The code for training and the networks (including SN) was an adapation of the code from the DDU \citep{mukhoti2023ddu} paper, which is available on github \url{https://github.com/omegafragger/DDU}.

\subsection{DDU Details}\label{appendix:ddu_details}
Let \(\bm{z}\) represent a feature vector, such as \(h(\bm{x})\). Feature densities are measured as,  
\begin{align}\label{eqn:feature_density}
    p(\bm z)&=\sum_{c} p(\bm z \mid y=c) p(y=c),
\end{align}
where $p(\bm{z} \mid y=c)$ represents a Gaussian evaluated at $\bm{z}$, with its mean and covariance estimated from all training instances labeled $c$, and $p(y=c)$ denotes the proportion of training instances with label $c$.

\textbf{Class-wise moments} for the Gaussians in \cref{eqn:feature_density} are computed as, 
\begin{align}\label{eqn:gda_eqns}
    \bm \mu_{c}&=\frac{1}{\mid \mathcal{I}_c \mid }\sum_{i\in \mathcal{I}_c} h(\bm x_i), \hspace{0.5cm} \\
    \bm \Sigma_{c}&=\frac{1}{\mid \mathcal{I}_c \mid - 1}\sum_{i\in \mathcal{I}_c} (h(\bm x_i) - \bm \mu_{c}) (h(\bm x_i) - \bm \mu_{c})^{T},
\end{align}

where $\mathcal{I}_c$ denotes the instances in the training set that have label $c$. 

\textbf{Intermediate Feature Densities:} Intermediate feature densities are computed by replacing the feature vector $h(\bm{x})$ with a projected and normalized $h^{(l)}(\bm{x})$. The projection and normalization are performed using \cref{alg:reshape_scale_project}, followed by an additional PCA-based projection to further reduce the dimensionality of the feature vector used in \cref{eqn:feature_density}.

\textbf{Intermediate Network Predictions:} Intermediate network predictions are generated by fitting a Bayesian multinomial logistic regression model to the feature vector produced by \cref{alg:reshape_scale_project}. A Laplace approximation is employed to estimate the posterior over the model weights. These predictions are denoted as "PSC" in the results presented in \cref{appendix:dirty_mnist}.

\section{Additional experiments and results}\label{appendix:additional_experiments}
In this section we present additional experimental results to supplement the work in \cref{sec:experiments}.

\subsection{Sensitivity and Smoothness Example}

Borrowing from \citet{amersfoot2020DUQ} we consider an example that demonstrates how the accuracy of the final prediction and the sensitivity of the model can be at odds. We additionally use this example to highlight that while SN helps ensure a smooth yet accurate classifier, as the penalty increases earlier layers become both sensitive and accurate.

In this example we consider $x_i\sim\mathcal{N}(0, \sigma^2)$ for $i = 1, 2$ and $y = \text{sign}(x_1) * \epsilon$, with $\epsilon$ denoting noise that flips the label on $y$ with a small probability. We generate 3000 and 1000 samples for training and validation, respectively. We set $\sigma = 0.3$ and the probability of a flip set to $0.001$. We then fit a neural network to predict the class label and we use a sequence of linear layers with Leaky-ReLU activations and hidden dimensions of size $[4,2,2,2,2]$. Additionally, we use residual connections for all the layers of shape $2$. The network is trained for 350 epochs using an Adam optimizer \citep{kingma2014adam} with the learning rate adjusted from 3e-2 to 3e-4 using cosine annealing \citep{loshchilov2016cosine}, additionally we set the weight-decay to 1e-5. 

During training, we apply the SN procedure from \citet{mukhoti2023ddu} using two different values for the SN constant: one large and one small. After training, we propagate the training data through the neural network and perform PCA on the intermediate representations. We then select a training instance and apply two perturbations to generate two new pseudo data points. The first perturbation is \textit{irrelevant}, as it modifies the input $\bm{x}$ along the second dimension, which does not affect the class label. The second perturbation flips the sign, thereby changing the class label.

\begin{figure*}
    \centering
    \includegraphics[width =1.0\textwidth]{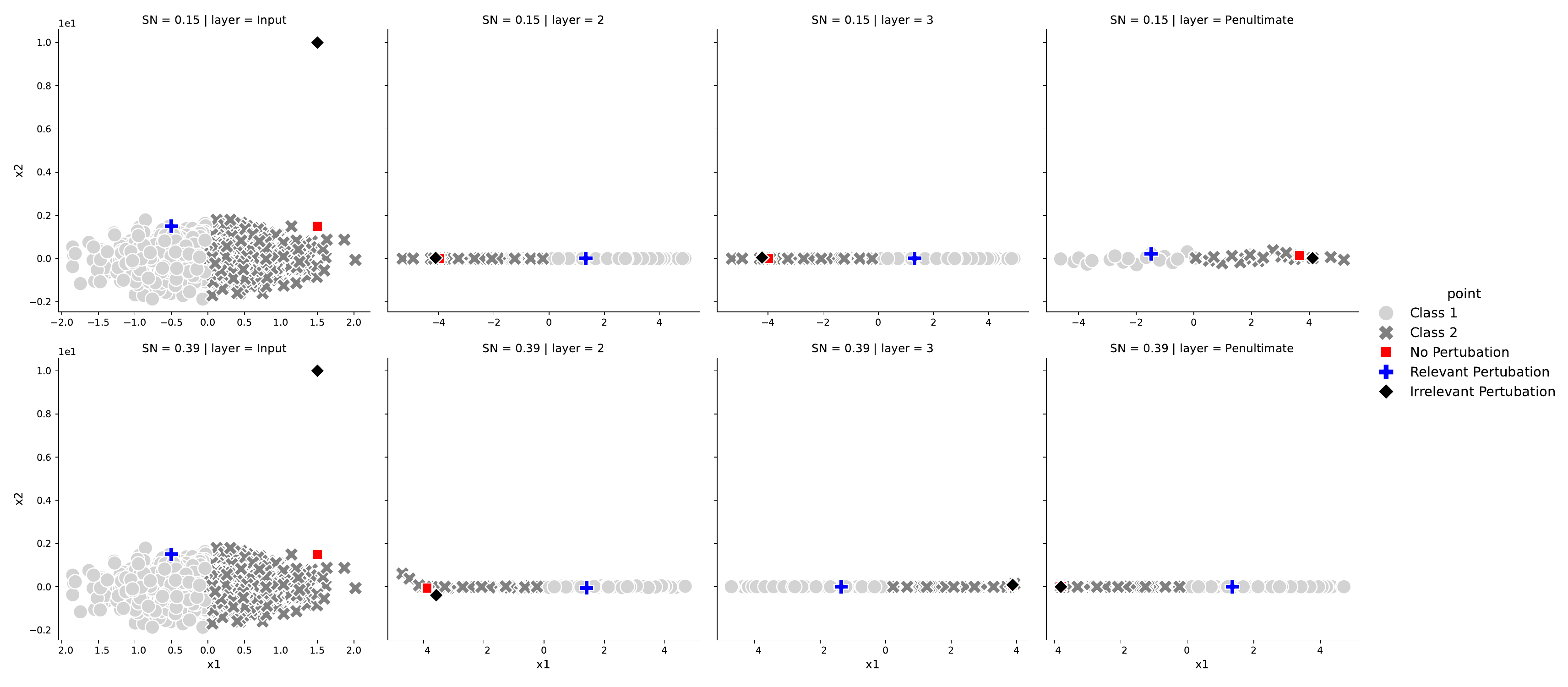}
    \caption{\textbf{As the upper bound for the spectral norm increases unique points collapse at earlier layers, but are unique for some early  layer.} This figure shows the 2D intermediate representations after PCA for the sign-example from \citet{amersfoot2020DUQ} as we change the upper bound for SN.}
    \label{fig:sign_example}
\end{figure*}

The results of this experiment are shown in \cref{fig:sign_example}. We see that for the smaller of the two spectral norm upper bounds that we can distinguish between the original input and the \textit{irrelevant} perturbation even at the penultimate layer. With the larger upper bound the penultimate layer appears to have collapsed the points. However, with the larger upper bound we see that one of the earlier layers still separates the two points but is collapsed by the next layer. Furthermore, we see that all layers correctly place the \textit{relevant} perturbation near the points of the opposite label. This shows that while the network seems to discard information necessary for sensitivity, SN ensures this information is preserved. If the spectral upper bound increases though, the layer where this information is preserved is earlier in the network. For deep networks, such as Resnet-50 in \cref{sec:feature_geometry_experiment}, the layer where this information is preserved is pushed earlier and earlier. Surprisingly, even without SN larger networks preserve this information at some earlier layer as demonstrated in the experiments in \cref{sec:experiments}.

\subsection{DDU Case Study}\label{appendix:dirty_mnist}
\begin{figure*}
    \centering
    \includegraphics[width=1.0\textwidth]{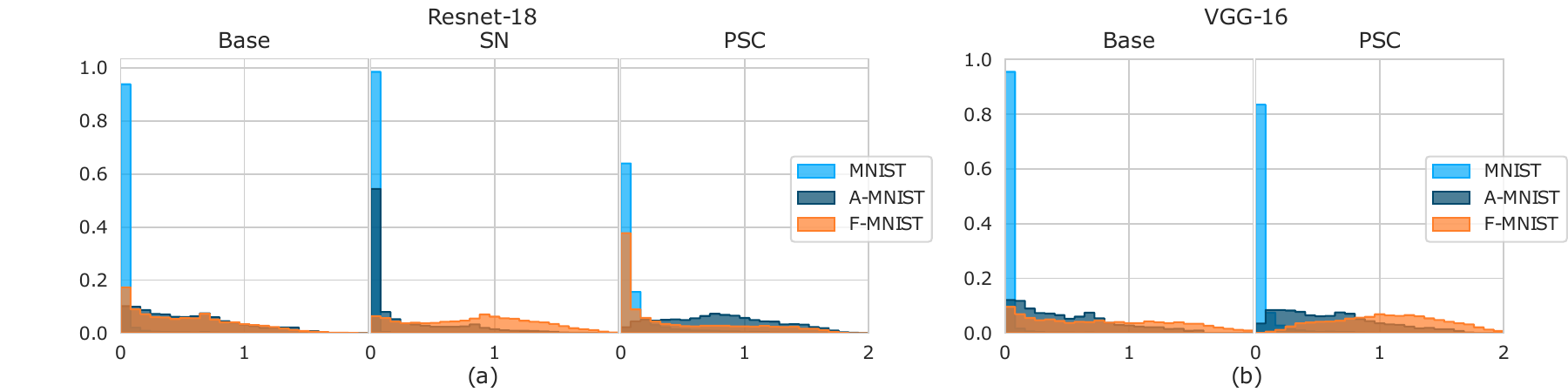}
    \caption{\textbf{Predictive entropy can be used to separate iD-Ambiguous and iD-Clean}. The figure shows the predictive entropy values when the base model used was Resnet-18 (a) or VGG-16 (b).}
    \label{fig:dirty_mnist_entropy}
\end{figure*}

\textbf{Predictive entropy} histograms from \cref{sec:ddu_case_study} are shown in \cref{fig:dirty_mnist_entropy}. In panel (b) we have the histograms for Resnet-18 and we see that the predictive entropy of the base model does well. Using SN doesn't improve on these results. Using the PSC pushes the entropy for iD-Ambiguous (A-MNIST) higher but also slightly increases the entropy for the iD-Clean (MNIST) examples. However, the result is we are able to separate the iD-Ambiguous from the iD-Clean with an AUROC that is similar to the base model. Combined with the feature densities of \cref{fig:dirty_mnist_density} we can distinguish all three data sources using the PSC. 

For VGG-16 in panel (b) of \cref{fig:dirty_mnist_entropy} we see a similar story as we did with Resnet-18. Using the PSC increases the predictive entropy of the iD-Ambiguous instances but also slightly increases the entropy of the iD-Clean instances.  However, the AUROC for using the PSC predictive entropy to separate iD-Clean and iD-Ambiguous is higher in this case. Just as before, we can combine these results with the feature density results to cleanly seperate all three data sources using the PSC. 

\end{document}